\documentclass[10pt, a4paper]{article}

\usepackage{subcaption}
\usepackage{adjustbox}

\usepackage[final]{lrec2026} 

\title{Forewarned is Forearmed:\\ When Non-sequential Embedding Turns Into an Anomaly Detector}

\name{Elys Allesiardo, Antoine Caubrière, Valentin Vielzeuf}

\address{Orange Research\\
         4 Rue du Clos Courtel, 35510 Cesson-Sévigné, France \\
         \texttt{firstname}.\texttt{lastname}@orange.com}

\abstract{
This paper offers an in-depth analysis of non-sequential multimodal sentence-level embeddings, with a particular focus on the SONAR model. We demonstrate that certain embedding dimensions are sensitive to perturbations and can serve as indicators of decoding anomalies. By leveraging the consistency between successive encoding and decoding, we successfully build an accurate detector. Additionally, we explore modifying specific dimensions of interest to attempt to correct them. This work underscores the importance of understanding and analyzing the embeddings themselves to enhance the reliability of multimodal representations.
 \\ \newline \Keywords{SONAR, multimodal, embeddings analysis, self-consistency} }

\begin{document}

\maketitleabstract

\section{Introduction}
Large Language Models (LLMs) have become the premier approach for addressing a large variety of tasks in Natural Language Processing, such as dialogue understanding, summary generation, or sophisticated reasoning~\cite{zhaosurvey}.
Moreover, this revolution has been extended to multimodal problems, with the rise of systems such as SpeechLLMs~\cite{ji2024wavchat,haque2019audio}, which combine a speech encoder and an LLM decoder in the hope to process speech contents with LLM's ability. These approaches are also permitted by the principle of encoder-decoder, often exploiting cross-modal attention~\cite{chan2016listen,chen2024bestow}
However, as observed by the community, processing long sequences of tokens is expensive and processing long continuous audio input even more~\cite{jia2025efficient,duquenne2023sonar}.

It is one reason why some recent works~\cite{duquenne2023sonar,khurana2022samu,feng2022language} have focused on the extraction of sentence-level encoder, from both text and audio inputs. These approaches consists in extracting a single non-sequential embedding of the given input (whether it is text or speech) and therefore allow efficient semantic similarity computation and (multilingual) information retrieval~\cite{khurana2022samu}. Moreover, these compact and non-sequential embeddings may also be decoded, as done by SONAR~\cite{duquenne2023sonar} and can help to build systems such as Large Concept Models (LCM)~\cite{barrault2024large} or SonarLLM~\cite{dragunov2025sonar}, by serving as a high-level concept tokenizer. 


These "concept" models are promising in the sense that they show improved performance on high-level tasks, such as summary generation, and can imply a considerable efficiency gain compared to vanilla LLMs~\cite{barrault2024large}. However, they heavily rely on the "concept" encoder. And it has been observed in~\cite{barrault2024large,duquenne2023sonar-expressive}, that Sonar may be prone to instability, propagating undesirable behaviors and creating anomalies in the output, and this even when using the based decoder of Sonar (infinite repetition of a random pattern, truncation of the transcribed sentence).

These anomalies may be compared to hallucinations observed for LLMs, which is a well-explored field of research. In particular, structural hallucinations~\cite{banerjee2025-structural-hallu} correspond to errors inherent in the generation process itself, such as repetitive loops, premature truncation, or degenerate sequence patterns. Detecting such anomalies remains challenging. Several 
approaches have been proposed, often relying on decoder-level signals or multiple generations.\cite{ma2025semanticenergydetectingllm} analyze log-probability distributions, showing that hallucinations exhibit distinct energy profiles;~\cite{sem_entropy} estimate semantic entropy by clustering multiple sampled generations; and~\cite{SelfCheckGPT} leverage self-consistency checks across diverse outputs. Although effective, these methods typically require access to internal decoding probabilities or incur high computational costs due to repeated sampling. 

For the specific case of a "concept" encoder such as SONAR, we propose another way to identify these anomalies\footnote{We choose to avoid the term hallucinations in this paper, as the definition is broad and the Sonar encoder-decoder model is not a chat model, targeting transcription or translation of a precise input.} by focusing on the embedding itself.
We summarize our contribution\footnote{https://github.com/Orange-OpenSource/sonar-speech-analysis} as follows \textbf{(a)} We first propose an in-depth dimensional analysis of the response of SONAR embeddings to input variations (acoustic characteristics, order of words, semantics). \textbf{(b)} We then investigate the impact of some specific dimensions on the decoding performance. \textbf{(c)} It helps us to get the intuition that the textual embedding of an anomaly transcript would be far from the original embedding that was wrongly decoded and, therefore, that an anomaly detector can be implemented simply by computing the distance between those two embeddings.
\textbf{(d)} And we finally show that detecting these anomalies and understanding the privileged impact of some dimensions of the embedding is an important brick in the way to correct them.

\section{Background and Notation}
\begin{figure}[hb]
    \centering
    \includegraphics[width=1.05\linewidth]{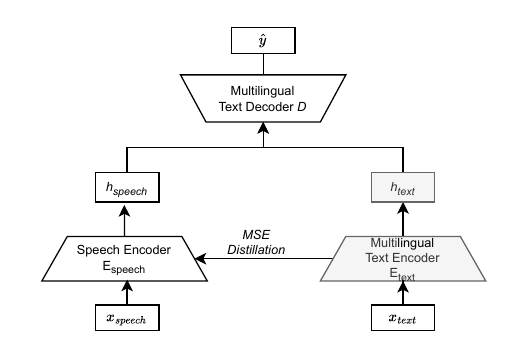}
    \caption{SONAR architecture}
    \label{fig:SONAR}
\end{figure}

Before diving into our analysis, we propose an overview of the SONAR model~\cite{duquenne2023sonar} and define the notation that will be used in the rest of the paper.
SONAR is a unified multi-modal and multi-lingual fixed-size sentence embedding space. As explained in the Introduction, this model is particularly well suited for tasks such as semantic similarity search across different languages or modalities, as well as machine translation, both text-to-text and speech-to-text. Automatic Speech Recognition (ASR) can also be performed (even in zero-shot scenarios) by encoding speech into the SONAR embedding space and then decoding it as text in the same language. We will focus especially on this decoding regime.

From Figure~\ref{fig:SONAR}, we can formalize the architecture.
The inputs $x_{speech}$ or $x_{text}$ are processed, respectively, by the speech encoder $E_{speech}$ or the text encoder $E_{text}$ to extract a non-sequential representation $h_{speech}$ or $h_{text}$.
Then, this embedding is fed to the decoder $D$ to generate a text $\hat{y}$, which may be transcription or a translation of the original input.
$E_{text}$ and $D$ are initialized from the encoder-decoder architecture of NLLB 1B~\cite{costa2022no}. This textual model is then finetuned on multiple tasks at the same time: translation, auto-encoding, denoising and cross-lingual similarity.
$E_{speech}$ is initialized from W2V-BERT 2.0~\cite{wa2V-bert2}, combined with a 3-layer attention pooling transformer, in order to achieve a non-sequential representation.
$E_{speech}$ is trained by MSE distillation between $h_{text}$ (extracted by the frozen $E_{text}$ from ground-truth text labels) and $h_{speech}$ (extracted from speech inputs).
At inference time, the transcription process may be formally described as:
\begin{equation}
\hat{y} = D(E_{speech}(x_{speech}))    
\end{equation}

All our experiments are conducted using the publicly available SONAR models. 
For tasks related to speech transcription and text encoding, we leverage the well-known LibriSpeech dataset~\cite{7178964}, specifically the test-clean subset.
By applying the pipeline, we achieve a Word Error Rate (WER) of 20.2 and a BERTScore~\cite{zhangbertscore} (F1) of 0.950.

\section{How do the representations respond to variations in the input?}



\subsection{Signal perturbations}
\textbf{Basic perturbations}
In this section, we assess the robustness of the SONAR embeddings to input perturbations.
We apply speed [0.5x;2x] (steps 0.25) and pitch shifts [-9;+9] (steps +3) to the raw input audio.
For each transformation, we compute the average embeddings at dataset-level and quantify its deviation from the baseline computed on ground-truth audio. 

\begin{figure}[ht]
    \centering
    \includegraphics[width=\linewidth]{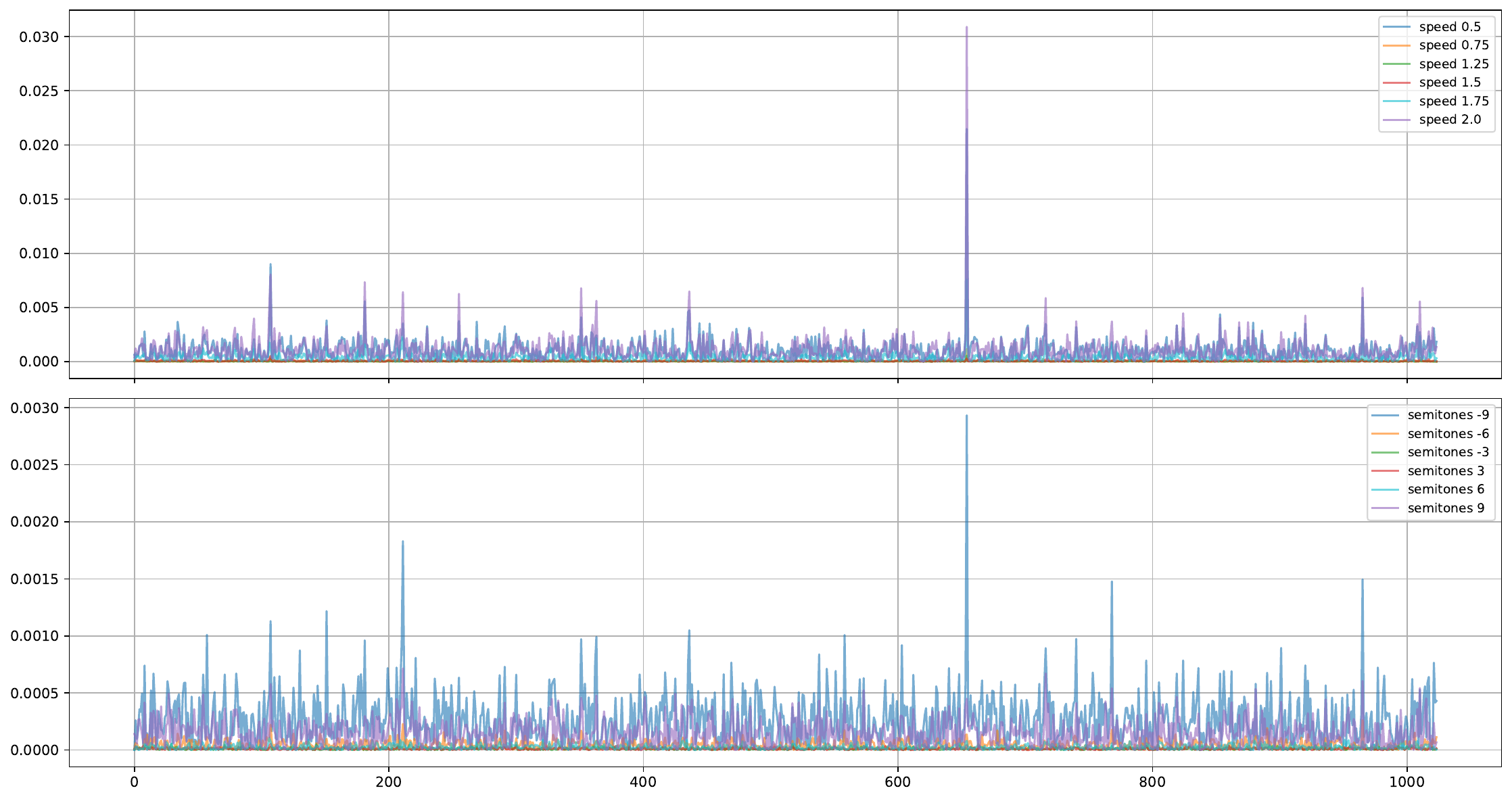}
    \caption{Absolute per-dimension deviation of SONAR embeddings under speed (top) and pitch (bottom) transformations.}
    \label{fig:audio_transform}
\end{figure}

For both types of perturbations, Figure~\ref{fig:audio_transform} shows a similar pattern: deviation increases rapidly over all dimensions under extreme setting (e.g. speed 0.5x or 2x ; pitch ±9 semitones) but are negligible for more realistic perturbations (e.g. speed 0.75x or 1.25x ; pitch ±3 semitones).
Since such extreme settings are unlikely for natural speech, these results suggest that SONAR embeddings are robust to small audio transformations, consistent with their semantic nature.

\textbf{Order perturbations}
We perturb the temporal order of the input signal by shuffling at word level, following the word boundaries obtained by automatic alignment using Montreal Forced Aligner (MFA)~\cite{mcauliffe2017montreal}\footnote{https://github.com/MontrealCorpusTools/Montreal-Forced-Aligner}.
We apply an n-gram shuffling procedure that consists of randomly permuting contiguous blocks of n words.
A higher rank shuffles larger contiguous blocks, so only subparts of the sentence are permuted, whereas rank 1 shuffles individual words. Results are reported in table ~\ref{tab:nGram}.

\begin{table}[h]
  \centering
  \begin{tabular}{|c|c|c|}
    \hline
    \textbf{nGram rank} & \textbf{WER} & \textbf{BERTScore (F1)} \\ \hline
    \textbf{6} & 46.3 & .905 \\ \hline
    \textbf{5} & 49.4 & .901 \\ \hline
    \textbf{4} & 55.0 & .885 \\ \hline
    \textbf{3} & 62.6 & .880 \\ \hline
    \textbf{2} & 73.7 & .857 \\ \hline
    \textbf{1} & 87.9 & .816 \\ \hline
  \end{tabular}
  \caption{Decoding performance depending on the shuffling rank.}
  \label{tab:nGram}
\end{table}

WER and BERTScore show that as the degree of semantic shuffling increases, the transcriptions tend to lose their semantic coherence.
As we reduce the rank, the decoder produces more anomalies, including artifacts (e.g., repetitions of words).
The system also tends to prioritize semantically plausible block of words over fidelity to the acoustic signal.
These results show that the input sequence order is not encoded within the embeddings and suggest that the encoded semantics drive the decoder in constructing the output order.

\textbf{Speaker and Pitch}
We perform speaker information probing experiments on the VoxCeleb dataset~\cite{Nagrani17}. The probing head is trained to simultaneously predict speaker identity and pitch information from speech SONAR embeddings. Our setup includes 1,251 unique speakers with balanced test sets (2 examples per speaker).
The probing architecture consists of a shared MLP backbone followed by task-specific heads. 
On the test set, the probing head achieves a speaker identification accuracy of 0.026 which, while being 33x higher than the random baseline, is still very low. For pitch prediction, the model fails to capture the true variations, as evidenced by a high median error ($\sim$79 Hz) and nearly 40\% of cases exceeding 100 Hz. This indicates that the system largely outputs values close to a global average rather than accurately tracking the ground-truth pitch. Thus, it confirms the weak relationship between speaker information and SONAR speech embeddings. 

\subsection{Impact of duration}
\label{sec:duration}
We investigate the effect of input sequence duration within the SONAR embedding space by comparing the embeddings produced by $E_{speech}$ and $E_{text}$.
To this end, we produce a synthetic corpus of random words.
We vary the sequence length from 1 to 40 by concatenating a variable number of words, each selected from a fixed vocabulary of 338 words.
To generate the speech version of the sentences, we utilize coqui-tts\footnote{https://github.com/coqui-ai/TTS} with the VITS Text-To-Speech model.
By applying the pipeline on this random spoken words corpus, we achieve a BERTScore (F1) of 0.813 for $E_{speech}$ embeddings and 0.884 for $E_{text}$ embeddings. 

For each of the 1024 dimensions, we analyze the monotonicity of the dimension values as a function of the sequence duration. 
\begin{figure}[ht]
    \centering
    \begin{subfigure}{\linewidth}
        \centering
        \includegraphics[height=0.21\textheight]{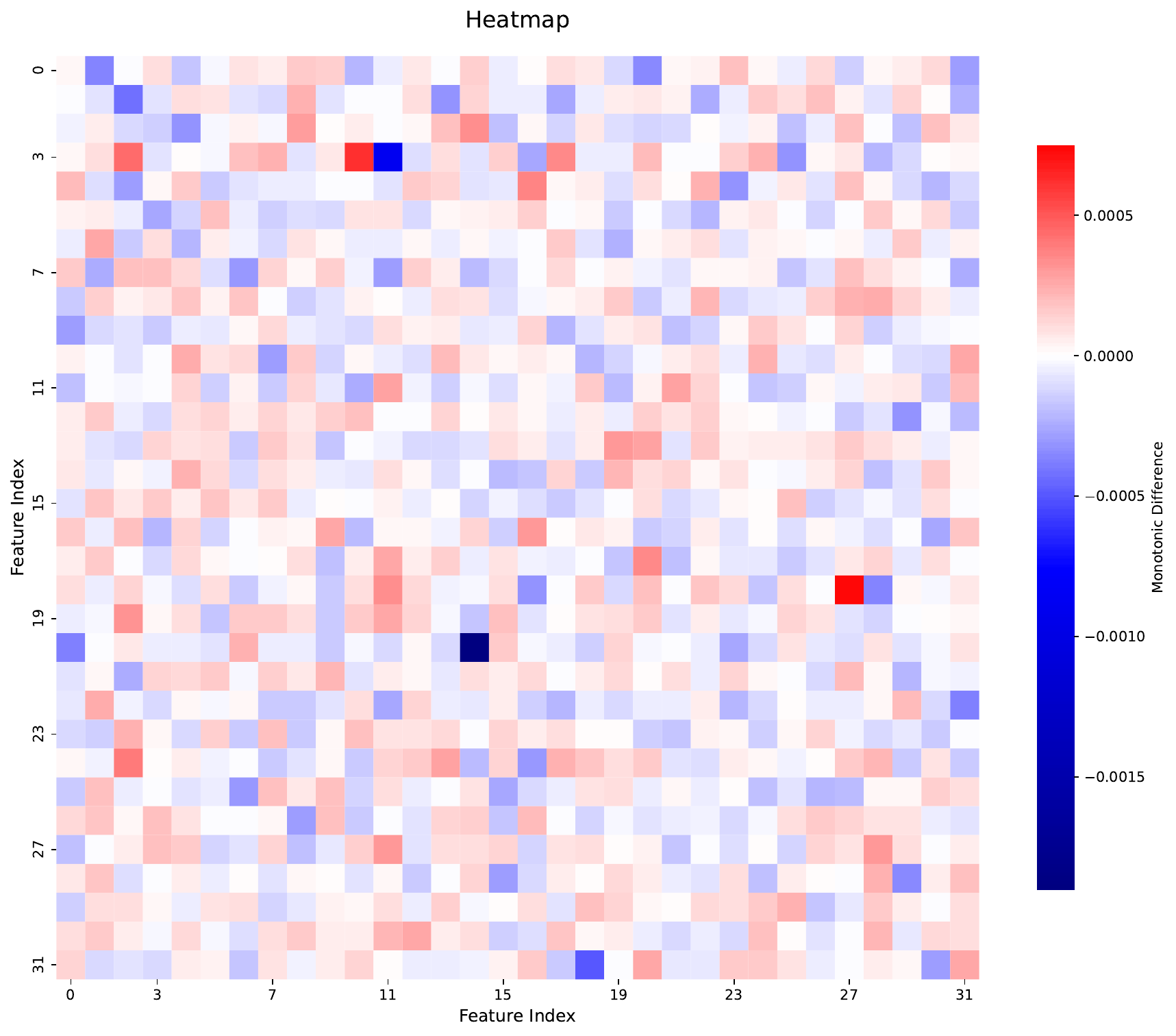}
        \caption{$E_{speech}$}
        \label{fig:speech_heatmap}
    \end{subfigure}
    \begin{subfigure}{\linewidth}
        \centering
        \includegraphics[height=0.21\textheight]{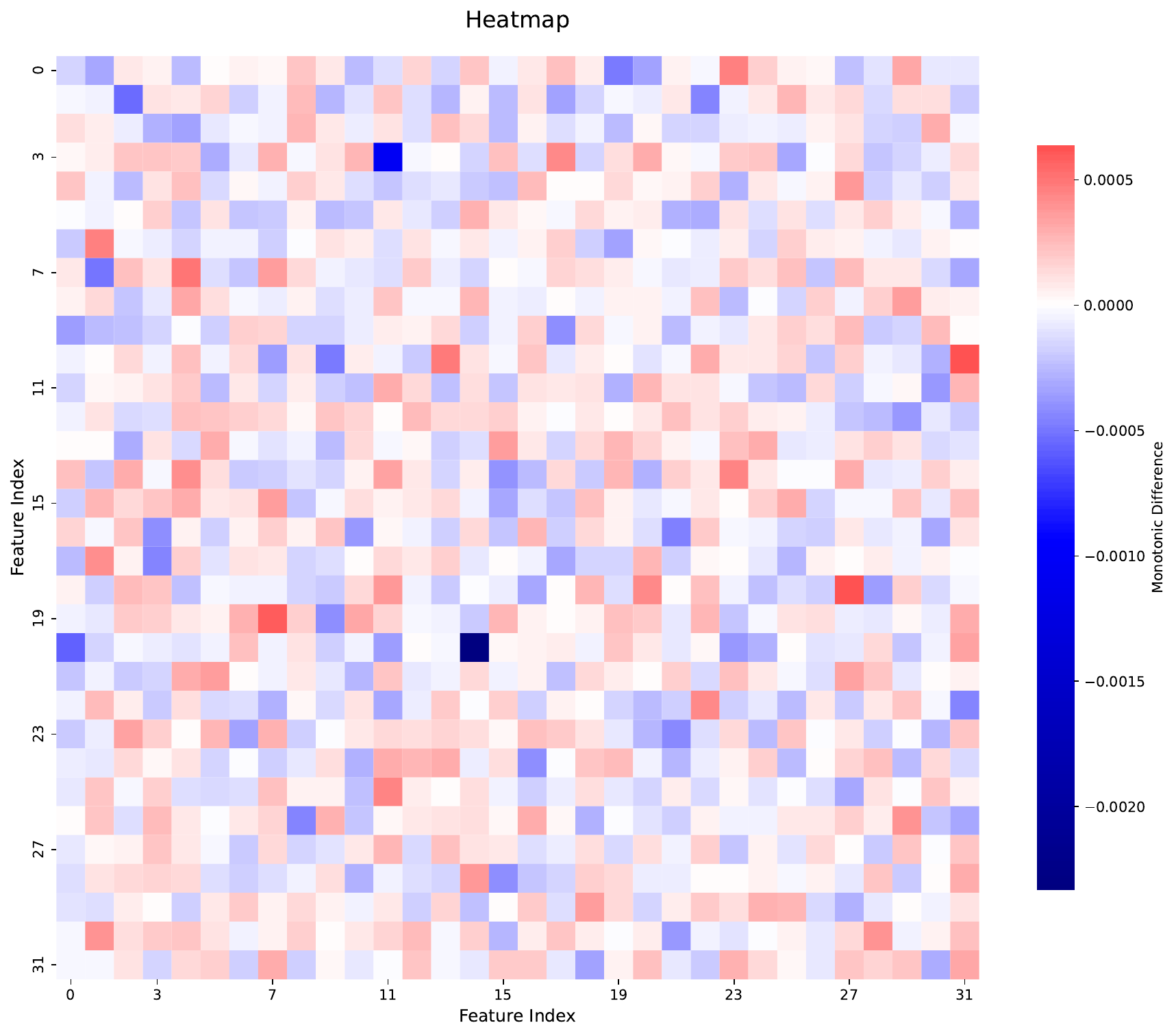}
        \caption{$E_{text}$}
        \label{fig:text_heatmap}
    \end{subfigure}
    \caption{32×32 monotonicity heatmaps. Each cell corresponds to a single embedding dimension.}
    \label{fig:text_speech_correlation}
\end{figure}

\begin{figure}[ht]
    \centering
    \includegraphics[width=\linewidth]{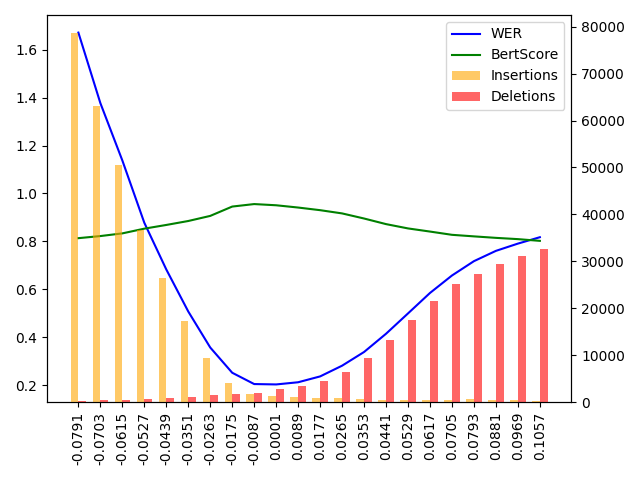}
    \caption{Effect of 654th‑dimension offsets on decoding performance.}
    \label{fig:ins_del-654}
\end{figure}

Figure~\ref{fig:text_speech_correlation} shows two 32x32 heatmaps, where each cell corresponds to the monotonicity of a dimension.
The left heatmap corresponds to the embeddings extracted from $E_{speech}$, while the right one corresponds to $E_{text}$.
These heatmaps show differences between $E_{speech}$ and $E_{text}$, although a few dimensions appear to maintain similar behavior. 
More specifically, dimensions 107 (l3, c11) and 654 (l20, c14)  seem to preserve a similar monotonicity. 
Such similarity between the two encoders suggests that these dimensions may encode information related to sequence length at these indices.
We investigate how the value of these dimensions affects the length of the output sequence, with a focus on the dimension 654, which shows the lowest monotonicity.
To assess the impact of this dimension, we perform a grid search over offsets applied to this single dimension across the entire corpus.
The grid search spans the observed minimum and maximum of this dimension in the dataset, with 22 steps.
For each step, we apply $D$ and compute the WER and BERTScore.
We also record the number of insertions and deletions in the automatic transcriptions.
The results are reported in Figure~\ref{fig:ins_del-654}.

We observe a correlation between this dimension's value and the decoder's output length.
Reducing this dimension constrains $D$ to produce words, whereas increasing it leads $D$ to discard most of the speech transcript.
A preliminary analysis of word insertions across different offsets suggests that most insertions are repetitions of words or word sequences already present in the sentence. 
We also denote the existence of this kind of undesirable behaviors even when no offsets are applied.

Finally, this section leads to develop the intuition that (a) variations in embeddings may be linked to undesirable behaviors, paving the way for a detection method exploiting both text and speech embeddings and (b) that
adjustments in dimensions such as 654 may lead to anomaly mitigation.

\begin{figure}[ht]
    \centering
    \includegraphics[width=0.9\linewidth]{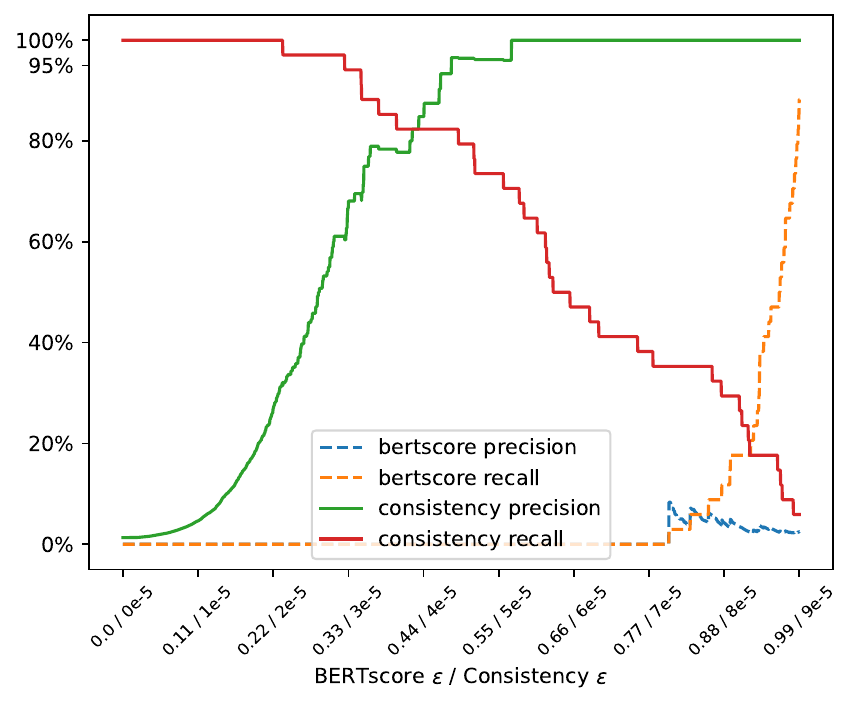}
    \caption{Precision and Recall of our anomaly detector and BERTScore baseline on the annotated LibriSpeech subset for various $\epsilon$.}
    \label{fig:precision_recall}
\end{figure}
\section{Undesirable behaviors}
\textbf{Detecting anomalies}

The SONAR embedding model may be prone to undesirable behaviors, reducing its usage as a backbone for larger applications (e.g. for Large Concept Models~\cite{barrault2024large}). Being able to automatically detect such anomalies is therefore of interest, as it may allow us to mitigate their impact on the rest of the chain.

To better our understanding of the kind of anomalies encountered, we extract the embeddings and transcriptions of the 2620 samples from the test-clean subset of LibriSpeech. 
Then we propose an annotation of all transcriptions\footnote{https://github.com/Orange-OpenSource/sonarspeech-analysis.}, consisting in a classification between four types of behavior: normal (e.g. usual transcription that may be paraphrased or not perfect), loop (e.g. repeating several times the same pattern of words), truncation (omitting a large part of the sentence), meaning (e.g. negating the original sentence).

Recall that embedding extraction consists of using $E_{speech}$ or $E_{text}$ to process a given input of the respective speech and text modalities. After getting the non-sequential embedding (let us denote it $h_1$ here), we are using the same decoder $D$ to generate a transcription or a translation $\hat{y}_1$. 
We propose using $E_{text}$ to encode $\hat{y}_1$ and obtain a second embedding $h_2$ (which could also be decoded with $D$ into $\hat{y}_2$).
Then, we compute the mean squared error between the two embeddings $H_1$ and $H_2$.
\begin{equation}
    MSE = d_{consistency} =\frac{1}{n}\sum_{i=1}^{n}(H_{2i} - H_{1i})^2.     
\end{equation}
In this specific case $n=1$, thus we simplify $d_{consistency} = (h_2-h_1)^2$ . This distance is a scalar and will be denoted as \textit{consistency} in the rest of the paper. We will consider that an anomaly is detected if the \textit{consistency} is greater than a fixed threshold $\epsilon$. 

We report in Figure~\ref{fig:precision_recall} the precision and recall of the designed detector for various $\epsilon$ choices. For the rest of the experiments, we choose to set $\epsilon=4.39e-5$.
We can compare our proposed approach with a naive baseline that consists of computing the WER or the BERTScore between $\hat{y}_1$ and $\hat{y}_2$. This method does not allow to detect anomaly at all, while the best precision/recall compromise of $d_{consistency}$ is 97\% / 82\%, enabling accurate anomaly detection. 

\begin{figure}[ht]
\noindent
\framebox[\linewidth][l]{%
  \begin{minipage}{0.98\linewidth}
\scriptsize{\textbf{$\hat{y}_1$} 

come back come back comeback comeback comeback comeback comeback comeback comeback comeback comeback comeback comeback comeback comeback comeback comeback comeback comeback comeback comeback comeback

\textbf{$\hat{y}_2$} 

come back come back comeback comeback comeback comeback comeback comeback comeback comeback comeback comeback comeback comeback comeback comeback comeback comeback comeback

BERTscore = 0.999, WER = 8\%, $d_{consistency}=8.96e-05$}
  \end{minipage}
  
}
\caption{Illustration of one loop anomaly from a sample of LibriSpeech test-clean. BERTscore and WER are not discriminative, while $d_{consistency}$ is greater than the anomaly threshold.}
\label{fig:anomaly_example}
\end{figure}
To explain the difference with the performance of the \textit{consistency}, we can focus on a specific example of anomaly in Figure~\ref{fig:anomaly_example}, encountered in LibriSpeech test-clean. This is a loop anomaly, repeating the term "come back" instead of the correct transcription. Both transcripts are very similar, and therefore it is not possible to predict from them that something wrong is happening. Using $d_{consistency}$ helps to avoid this problem. An intuition that can explain this ability is that the embedding $h_2$ can be very different from $h_1$, simply because $\hat{y}_1$, during an anomaly, is a rarely encountered input by the text encoder and therefore particularly difficult to encode.

\begin{figure}[hb]
    \centering
    \includegraphics[width=\linewidth]{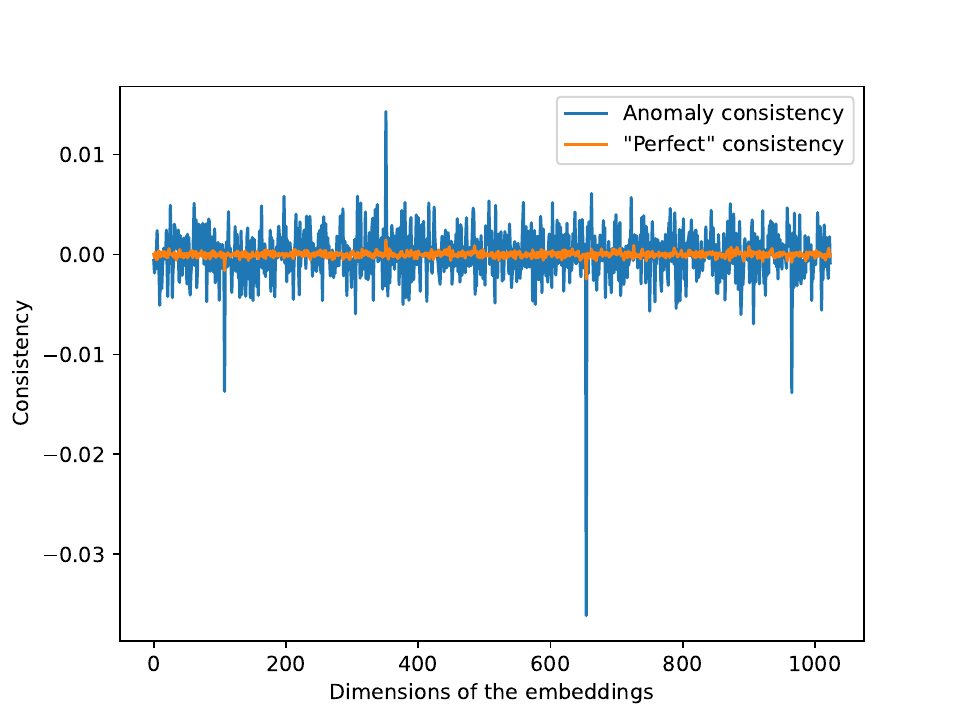}
    \caption{Consistency for two different categories: anomaly and "perfect". Anomaly consistency is computed on the average of all embeddings leading to anomalies. "Perfect" consistency is computed on the average of embeddings leading to a perfect transcript.}
    \label{fig:embedding_diff}
\end{figure}

Figure~\ref{fig:embedding_diff} may help to confirm this intuition. We observe the average consistency of the anomalies compared to that of perfectly transcribed samples. The consistency of anomalies shows that there is a great gap in some specific dimensions of the embedding, for example, 654. It therefore confirms what has been described in previous sections: certain dimensions of the embeddings have a stronger impact on the decoding. And we observe here that they are changing a lot between $h_1$ and $h_2$, explaining the high $d_{consistency}$ and the anomaly detection process.

\textbf{Does a remedy to these anomalies exist?}

As we propose an efficient and accurate anomaly detector, a straightforward usage would be to combine it with an anomaly handler/corrector.
A first basic way for critical application would be to label the output with a warning, when an anomaly is detected.
Yet, another hypothesis that may need exploration is that the decoder's hyperparameters are not adapted for the inputs conducting to anomalies, and therefore, changing these very parameters would help reducing undesirable outputs.
Thus we select the subset of LibriSpeech test-clean containing anomalies (which is a small dataset of around 50 samples) and we propose to perform a large grid search on the decoder hyper-parameters (temperature, length penalty, and beam size). We evaluate all configurations in terms of BERTScore. We observe no significant improvement on this subset compared to using the default configuration. 

As decoder parameters do not improve anomalies, we investigate applying controlled perturbation to embeddings.
By examining consistency difference between 'perfect' and anomaly embeddings (Figure~\ref{fig:embedding_diff}), we identify two dimensions that deviate the most (654,351).
We include the two dimensions that deviate the least (695, 932) and two randomly chosen dimensions (17, 666).
To find optimal values of these dimensions, we perform an offset grid search on the anomaly subset using 22 steps. As in Section~\ref{sec:duration}, this spans the observed range of dimensions in the test-clean dataset.
We then compute the mean consistency in Table~\ref{tab:grid_hallu_concist}. We observe that overall, the grid search allows to find values with a positive impact on consistency, compared to the baseline. Moreover, the more dimensions are prone to deviate, the greater their impact on consistency. Yet, for the specific case of loop anomalies, the opposite behavior is observed, with dimensions 932 and 695 obtaining the best consistency. 

\begin{table}[h]
\begin{adjustbox}{max width=\linewidth}
\centering
\begin{tabular}{|lcccc|}
\hline
 & \textbf{meaning} & \textbf{loop} & \textbf{truncate} & \textbf{all} \\ \hline
 \textit{baseline}  & 11.98 & 8.889 & 5.991 &  6.530 \\
 \hline
 2 most  & 2.918 & 4.525 & 3.730 & 3.977 \\
 2 least & 6.551 & 3.904 & 4.540 & 4.727 \\
 2 random & 3.540 & 5.142 & 3.971 & 4.375 \\
\hline
\end{tabular}
\end{adjustbox}
\caption{Best mean consistency obtained through grid search. The baseline corresponds to no offset applied. Numbers are scaled by 1e5 for readability.}
\label{tab:grid_hallu_concist}

\end{table}

These first results confirm that modifying specific dimensions when trying to limit anomalies will have a greater impact, paving the way towards an efficient anomaly correction method.
However, thinking dimensions are independent would be inaccurate, and more general approaches, notably focusing on modifying the training criterion of the SONAR distillation, may be promising.



\section{Conclusion}
We conducted an in-depth analysis of non-sequential embeddings, highlighting strong variations in specific dimensions associated with input perturbations. Based on these insights, we developed a lightweight and accurate anomaly detection method resting on the idea that the SONAR embedding of the decoded text should be close to the first extracted embedding. We also proved that modifying only specific well-targeted dimensions of the embedding may prevent some of the anomalies.
Future work may explore trainable remedies to the embedding,
for instance, changing the distillation paradigm of SONAR by switching to cross-modal alignment and adding consistency as an auxiliary criterion.


\section{Bibliographical References}\label{sec:reference}

\bibliographystyle{lrec2026-natbib}
\bibliography{lrec2026-example}
\end{document}